\documentclass[journal]{IEEEtran}
\usepackage{times}
\usepackage{graphicx}
\usepackage{bbm}
\usepackage{subfigure}
\usepackage{algorithm}
\usepackage{algorithmic}
\usepackage{multicol}
\usepackage{multirow}
\usepackage{booktabs}
\usepackage{array}
\usepackage[T1]{fontenc}
\usepackage[latin9]{inputenc}
\usepackage{cite}
\usepackage{color}
\usepackage{collcell}
\usepackage{hhline}
\usepackage{pgf}
\usepackage{pgfplots}
\usepackage{pdfpages}
\usepackage{amssymb}
\usepackage{amsmath,bm}
\usepackage{amsfonts}
\usepackage[colorlinks, linkcolor=red, anchorcolor=blue, citecolor=green]{hyperref}
\usepackage{graphicx}
\usepackage{rotating}
\usepackage{changepage}
\usepackage{makecell}
\usepackage{color, colortbl}
\usepackage{url}
\usepackage{pifont}
\usepackage{xcolor}
\usepackage{threeparttable}

\newcommand\Tstrut{\rule{0pt}{2.6ex}}         
\newcommand\Bstrut{\rule[-0.9ex]{0pt}{0pt}}   

\ifCLASSINFOpdf
\else
\fi
\begin{document}


\title{FuTH-Net: Fusing Temporal Relations and Holistic Features for Aerial Video Classification}

\author{Pu~Jin,
    \and Lichao~Mou,
    \and Yuansheng~Hua,
    \and Gui-Song~Xia, 
    \and  Xiao~Xiang~Zhu

\thanks{This work is jointly supported by the European Research Council (ERC) under the European Union's Horizon 2020 research and innovation programme (grant agreement No. [ERC-2016-StG-714087], Acronym: \textit{So2Sat}), by the Helmholtz Association through the Framework of Helmholtz AI (grant  number:  ZT-I-PF-5-01) - Local Unit ``Munich Unit @Aeronautics, Space and Transport (MASTr)'' and Helmholtz Excellent Professorship ``Data Science in Earth Observation - Big Data Fusion for Urban Research''(grant number: W2-W3-100), by the German Federal Ministry of Education and Research (BMBF) in the framework of the international future AI lab ``AI4EO -- Artificial Intelligence for Earth Observation: Reasoning, Uncertainties, Ethics and Beyond'' (grant number: 01DD20001) and by German Federal Ministry of Economics and Technology in the framework of the ``national center of excellence ML4Earth'' (grant number: 50EE2201C). 

P. Jin is with the State Key Laboratory for Information Engineering in Surveying, Mapping and Remote Sensing (LIESMARS), Wuhan University, Wuhan 430072, China, and also with the Department of Aerospace and Geodesy, Technical University of Munich (TUM), Germany (e-mail: pu.jin@tum.de).

L. Mou, Y. Hua, and X. X. Zhu are with the Remote Sensing Technology Institute, German Aerospace Center, 82234 We{\ss}ling, Germany, and also with the Data Science in Earth Observation (former: Signal Processing in Earth Observation), Technical University of Munich, 80333 Munich, Germany. (e-mails: lichao.mou@dlr.de; yuansheng.hua@dlr.de;  xiaoxiang.zhu@dlr.de).

G.-S. Xia is with the State Key Laboratory for Information Engineering in Surveying, Mapping and Remote
Sensing (LIESMARS), and also with the School of Computer Science, Wuhan University,
Wuhan 430072, China (e-mail: guisong.xia@whu.edu.cn).}

}

%
%

\markboth{}%
{Shell \MakeLowercase{\textit{et al.}}: Bare Demo of IEEEtran.cls for Journals}
%



\maketitle

\begin{abstract}
\textcolor{blue}{This work has been accepted by IEEE TGRS for publication.} Unmanned aerial vehicles (UAVs) are now widely applied to data acquisition due to its low cost and fast mobility. With the increasing volume of aerial videos, the demand for automatically parsing these videos is surging. To achieve this, current researches mainly focus on extracting a holistic feature with convolutions along both spatial and temporal dimensions. However, these methods are limited by small temporal receptive fields and cannot adequately capture long-term temporal dependencies which are important for describing complicated dynamics. In this paper, we propose a novel deep neural network, termed FuTH-Net, to model not only holistic features, but also temporal relations for aerial video classification. Furthermore, the holistic features are refined by the multi-scale temporal relations in a novel fusion module for yielding more discriminative video representations. More specially, FuTH-Net employs a two-pathway architecture: (1) a holistic representation pathway to learn a general feature of both frame appearances and short-term temporal variations and (2) a temporal relation pathway to capture multi-scale temporal relations across arbitrary frames, providing long-term temporal dependencies. Afterwards, a novel fusion module is proposed to spatiotemporal integrate the two features learned from the two pathways. Our model is evaluated on two aerial video classification datasets, ERA and Drone-Action, and achieves the state-of-the-art results. This demonstrates its effectiveness and good generalization capacity across different recognition tasks (event classification and human action recognition). To facilitate further research, we release the code at \url{https://gitlab.lrz.de/ai4eo/reasoning/futh-net}.
\end{abstract}


%
\IEEEpeerreviewmaketitle

\section{Introduction}
%
%
%
%

\IEEEPARstart{B}{y} the virtue of low-cost, real-time, and high-resolution data acquisition capacity, unmanned aerial vehicles (UAVs) can be exploited for a wide range of applications~\cite{pajares2015overview, xiang2019mini, zhu2017deep, colomina2014unmanned, tijtgat2017embedded,  teutsch2012detection, zhang2005object, xuan2019object, li2018r, dobrokhodov2006vision, wang2016detecting,  puri2005survey, kanistras2013survey, puri2007statistical,feng2015uav, everaerts2008use, rango2009unmanned} in the field of remote sensing, such as object tracking and surveillance~\cite{tijtgat2017embedded, teutsch2012detection, zhang2005object, xuan2019object, li2018r, dobrokhodov2006vision}, traffic flow monitoring~\cite{wang2016detecting, puri2005survey, kanistras2013survey, puri2007statistical}, and precision agriculture~\cite{feng2015uav, everaerts2008use, rango2009unmanned}. With the proliferation of UAVs worldwide, the number of produced aerial videos is significantly increasing. Hence there is an escalating demand for automatically parsing aerial videos, because it is unrealistic for humans to screen such big data and understand their contents.  Therefore, aerial video classification becomes an important task in aerial video interpretation~\cite{mou2020era}.

Feature learning and representation from videos is crucial for this task. Convolutional neural networks (CNNs) have demonstrated the superb capability of learning effective visual representations from images. For instance, ResNet~\cite{he2016deep} has achieved an impressive performance on the ImageNet dataset, which is even better than the reported human-level performance~\cite{ranjan2018deep}. Compared to a sequence of remote sensing images in which the temporal information is limited due to relatively long satellite revisit periods, an overhead video is able to deliver more fine-grained temporal dynamics that are essential for describing complex events. Therefore, moving from image recognition to video classification, much effort has been made to learning spatiotemporal feature representations.

On the one hand, several methods~\cite{KarpathyCVPR14, donahue2015long, yue2015beyond, simonyan2014two, qiu2017learningdeep, laptev2008learning, niebles2008unsupervised, wang2013action, wang2016actions, feichtenhofer2016convolutional,tran2015learning, ji20123d, taylor2010convolutional,  varol2017long,carreira2017quo, qiu2017learning, tran2018closer, tran2019video} aim at learning a global spatiotemporal feature representation that can holistically represent a video. A straightforward idea is to extract spatiotemporal features on each video frame individually by making use of 2D convolutions and then pool stacked feature maps across the temporal domain~\cite{KarpathyCVPR14}. However, this could lead to the ignorance of temporal relations among various frames. To address this,~\cite{donahue2015long} and~\cite{yue2015beyond} employ recurrent neural networks (RNNs) such as long short-term memory (LSTM)~\cite{hochreiter1997long} to model temporal relations by integrating features over time. But the effectiveness of such methods usually depends heavily on the learning effect of long-term memorization. Furthermore, 3D CNNs are fairly natural models for video representation learning and able to learn global spatiotemporal features by performing 3D convolutions in both spatial and temporal dimensions. Some 3D CNN architectures~\cite{tran2015learning, ji20123d, taylor2010convolutional,  varol2017long,carreira2017quo, qiu2017learning, tran2018closer, tran2019video} have been investigated and shown impressive performance. For instance, in~\cite{tran2015learning}, the authors propose a 3D CNN model with 3$\times$3$\times$3 convolution filters for learning a video representation on a large-scale video dataset. Nonetheless, massive computational consumption and memory demand hinder efforts to train a very deep 3D CNN, and limit the performance of 3D CNN architectures. To address this problem, inflated 3D convolution filters\cite{carreira2017quo} and decomposed 3D convolution filters~\cite{qiu2017learning, tran2018closer} utilize a more economic method to implement 3D convolutions and boost the performance of 3D CNNs. However, the aforementioned methods with either 2D or 3D convolutions have limited temporal receptive fields and therefore cannot adequately capture variable temporal dependencies. On the other hand, a few recent works attempt to explicitly model temporal relationships and demonstrate promising results in several tasks, to name a few, temporal relational reasoning\cite{lin2019tsm, mou2016spatiotemporal, liu2019learning, zhou2018temporal, wang2018temporal}, object detection and tracking\cite{teutsch2012detection, zhang2005object, xuan2019object, li2018r}, event recognition\cite{shu2015joint,feichtenhofer2019slowfast, wu2020multigrid}, video segmentation\cite{lyu2018uavid, mou2018vehicle, lyu2019lip}, dynamic texture recognition\cite{yang2016dynamic}, and spatiotemporal learning~\cite{gao2019pcc, li2021abssnet}.


A video delivers not only spatial information but also temporal dynamics. Hence, some studies are dedicated to capture spatial (appearance) and temporal (motion) representations separately by a two-stream architecture. In these two-stream models, fusing the features from two pathways is an important procedure for recognition. For example,~\cite{simonyan2014two} directly fuses the softmax scores using either averaging operation or a simple linear SVM. In~\cite{karpathy2014large}, the authors utilize a fully connected layer to merge the two streams of the late fusion model. However, its performance is surpassed by a purely spatial network. Additionally,~\cite{Feichtenhofer2016spatio} introduces residual connections between appearance and temporal streams to enable motion interactions. For stream fusion, the authors average the prediction scores of the classification layers from two streams. In~\cite{feichtenhofer2016convolutional}, the authors investigate several fusion methods such as max, concatenation, convolution, and observe that 3D convolutional fusion outperforms averaging the softmax output. The main limitation of the two-stream architecture is that it is not capable to spatiotemporally match spatial and temporal information. Therefore, a fusion method is needed to spatiotemporally register the features from two pathways. However, the abovementioned fusion methods leverage a single operation (e.g., averaging) that is not able to effectively enable spatiotemporal interactions between them.

The motion in aerial videos usually has different durations and shows a high variability. For example, in the ERA dataset~\cite{mou2020era}, \emph{mudslide} shows a simple and repeated motion over a long duration, which could be described by a few video frames; \emph{car racing} depicts a complicated, dynamic process and is composed of a variety of consecutive motions including chasing, approaching, away, colliding, etc. over a short duration. Temporal relations across multiple frames are an important cue to represent the complex motion. The aforementioned approaches based on spatiotemporal convolutions (e.g., $3\times 3\times 3$ convolutions) simply add a temporal dimension to 2D convolution filters to implicitly learn temporal dependencies, and they are not adaptable to capture various, complicated temporal dynamics over a long duration due to their limited temporal receptive fields. To address this issue, we propose to explicitly learn temporal relations across arbitrary frames to effectively model long-term temporal dependencies. Furthermore, we introduce multi-scale temporal relations into holistic features to design a two-pathway architecture for aerial video classification. Besides, for spatiotemporal registering temporal relations and holistic features, we propose a novel fusion module in which holistic features are spatiotemporally modulated with temporal relations.

\begin{figure*}[h]
	\centering
 	\includegraphics[width = 1\linewidth]{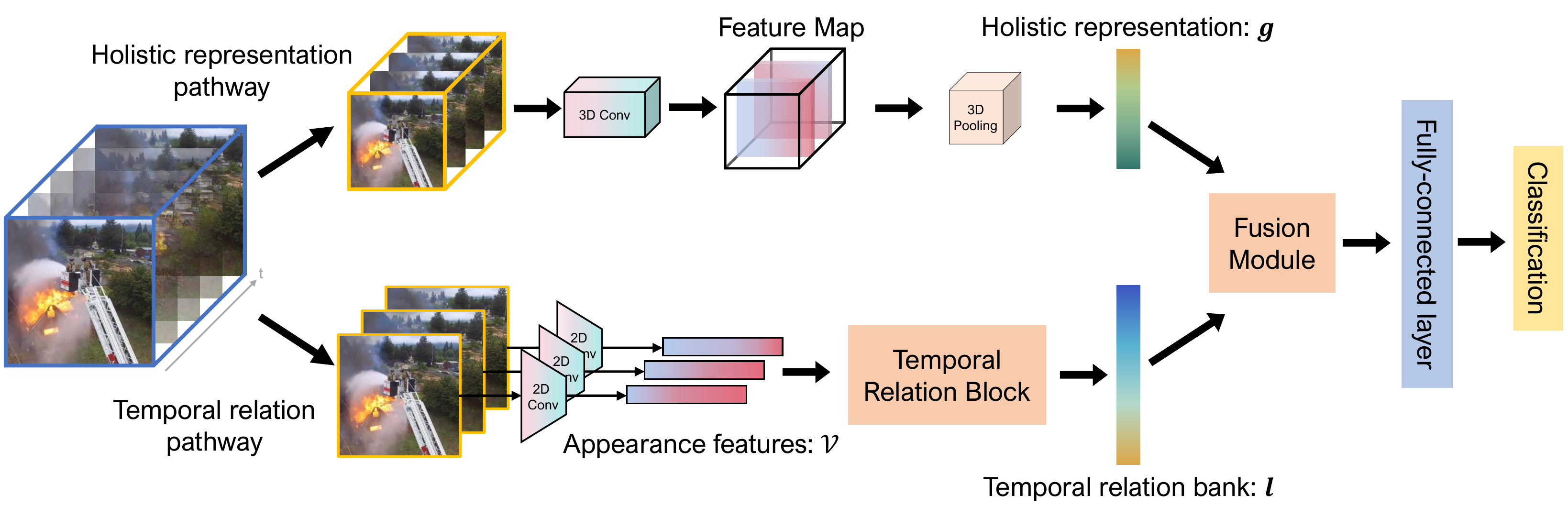}
	\caption{\textbf{The overview of FuTH-Net:} (1) The upper pathway, namely the holistic representation pathway, aims at capturing a holistic feature $\bm{g}$ by 3D convolutions. (2) The lower pathway, namely temporal relation pathway, aims to learn a multi-scale temporal relation bank $\bm{l}$ by a temporal relation block. (3) A followed fusion module combines the outputs of two pathways to generate a robust fused feature $\bm{z}$ which is finally fed into a fully-connected layer for aerial video classification.}
	\label{golfnet}
\end{figure*}

In this paper, we present a two-pathway network, termed FuTH-Net (\textbf{Fu}sing \textbf{T}emporal relations and \textbf{H}olistic features for aerial video classification). One pathway is devised to capture a holistic feature describing appearances and short-term temporal variations. The other pathway is responsible for excavating temporal relations across arbitrary frames at multiple timescales, providing long-term temporal dependencies. Last but not least, for spatiotemporally fusing two features from two pathways, we further present a novel fusion module in which the multi-scale temporal relations are leveraged to refine the temporal features in the holistic representation.  More specifically, we learn the holistic feature by treating a video as an entirety and using inflated 3D convolution operators~\cite{carreira2017quo}. Meanwhile, we sample frame-level feature vectors at different sampling rates to learn multi-scale temporal relations with a sequence of multilayer perceptrons (MLPs)~\cite{gardner1998artificial}. As to the fusion of these two features, we employ a fusion module in which the temporal relations are modulated with the holistic representation by a normalization-like process~\cite{ioffe2015batch, su2020adapting}. The resulting feature representation is then fed into the following layers for the purpose of video classification. Contributions of this paper are threefold:

\begin{itemize}
	\item We propose a novel network, namely FuTH-Net, for the task of aerial video classification. This network exploits a two-pathway architecture, one for learning a video presentation holistically and the other for fully excavating useful temporal relations at multiple timescales among video frames.

    \item A novel fusion module exploits a normalization-like pipeline in which the two features learned from two pathways are spatiotemporally registered by modulating the holistic features according to temporal relations. In this module, the temporal information in holistic features is refined by multi-scale temporal relations. A more discriminative fused feature is obtained for distinguishing different video events.
    
    \item We evaluate the effectiveness of the proposed network through extensive experiments, and experimental results show that our method achieves the state-of-the-art performance.
    
\end{itemize}

The remaining sections of this paper are organized as follows. Section \ref{II} details the architecture of FuTH-Net, and Section \ref{III} shows and discusses experimental results. The conclusion is drawn in Section \ref{IV}.

\section{Network Architecture} \label{II}

In this section, we detail our proposed network architecture, FuTH-Net, for aerial video classification. First, we introduce an overview of the proposed network in Section~\ref{A}. Furthermore, we give more detailed descriptions for two modules, temporal relation block and fusion module, in Section~\ref{B} and~\ref{C}. Finally, the implementation of our network is introduced in Section~\ref{D}.

\subsection{FuTH-Net} \label{A}

\begin{figure}[h]
	\centering
 	\includegraphics[width = 1\linewidth]{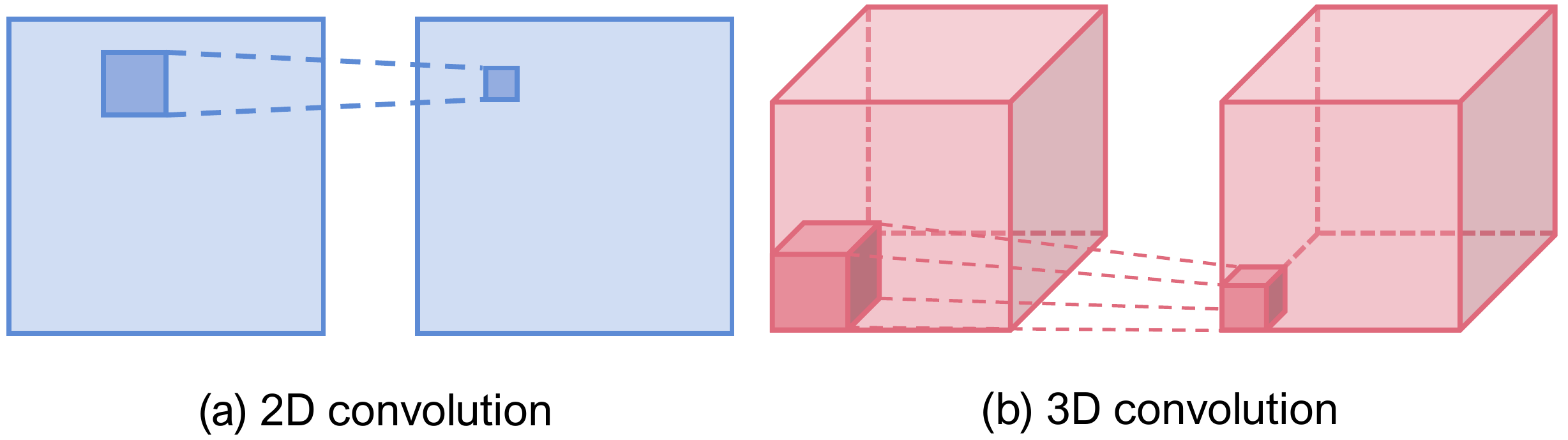}
	\caption{\textbf{2D convolution \emph{vs.} 3D convolution. Compared to 2D convolution, 3D convolution slides in both temporal and spatial dimensions and results in an output volume, which thereby captures both spatial and temporal information, i.e., the holistic representation.}}
	\label{3D_conv}
\end{figure}

The motivation of our network is to simultaneously model the holistic feature and temporal relations of a video with a two-pathway architecture. The resulting two feature representations are integrated by a fusion module. The overview of the architecture is illustrated in Fig.~\ref{golfnet}.

\textbf{Holistic representation pathway} treats a video as an entity and aims at learning a holistic feature by 3D convolutions. 3D convolution is achieved by endowing 2D convolution with an additional dimension (e.g., the temporal dimension of aerial videos), which is illustrated in Fig.~\ref{3D_conv}. Compared to 2D convolution, 3D convolution is able to capture both spatial and temporal information, so called holistic representation in our case. It is of importance for video classification under some circumstances where events with simple temporal dynamics are strongly associated with certain objects or scenes. As to the implementation of 3D convolutions, many efforts, e.g., 3D convolutional kernel~\cite{tran2015learning}, inflated 3D convolution~\cite{carreira2017quo}, and pseudo 3D convolution~\cite{qiu2017learning}, have been made to symmetrically extract both spatial and short-term temporal information. In this work, we choose a typical 2D CNN architecture and transform all 2D operations to 3D operations by a specific 3D implementation method~\cite{carreira2017quo}. Then, we employ the transformed 3D CNN with a bunch of 3D convolution and pooling operations on a video volume to capture a holistic representation $\bm{g}$.

\begin{figure}[h]
	\centering
 	\includegraphics[width = 1\linewidth]{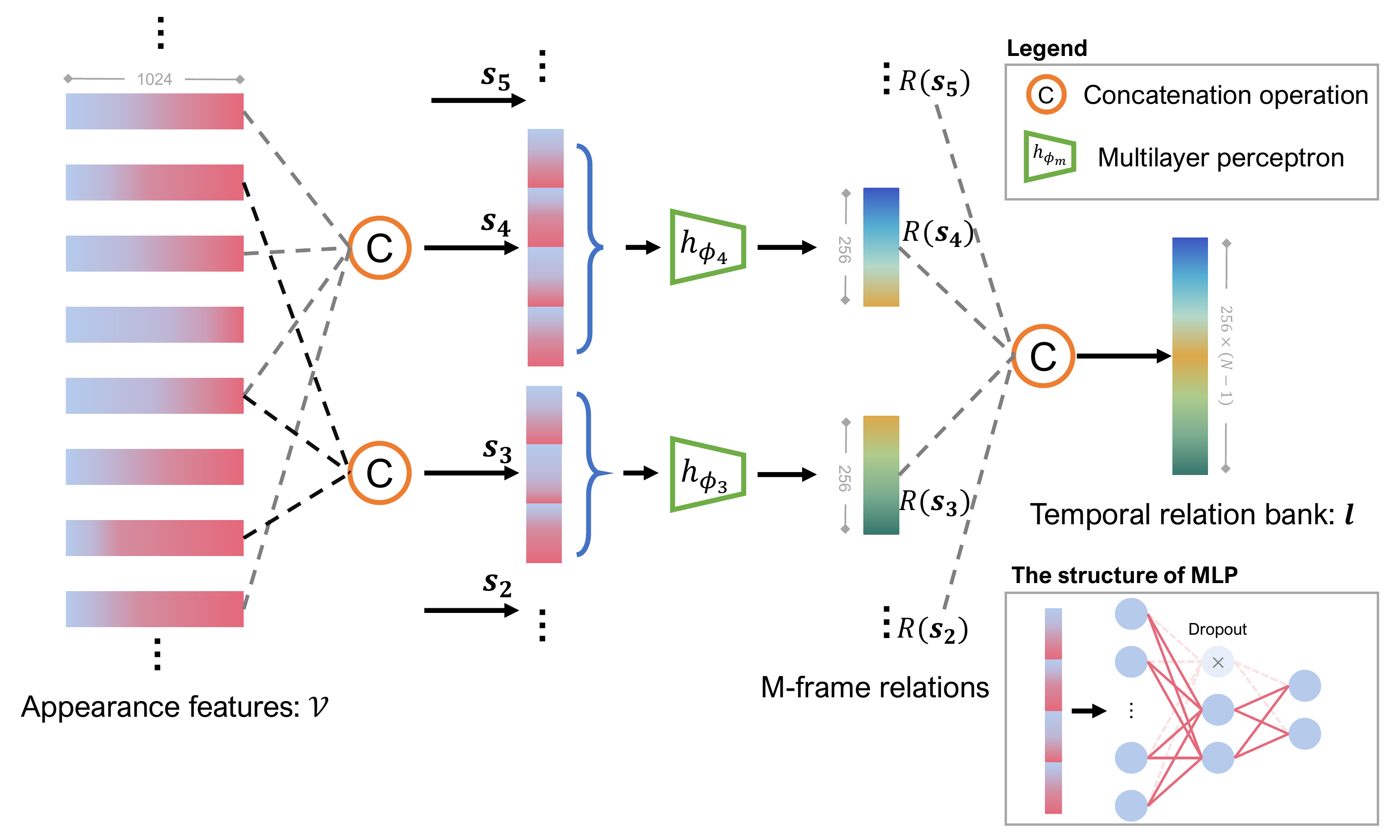}
	\caption{\textbf{Temporal relation block.} Appearance feature vectors are randomly selected from the feature set $\mathcal{V}$. Afterwards, the selected vectors are concatenated and then fed into an MLP to learn a corresponding \emph{m-frame} relation. Finally, all \emph{m-frame} relations are concatenated to produce a multi-scale temporal relation bank $\bm l$.}
	\label{trb}
\end{figure}

\textbf{Temporal relation pathway} views a video as a sequence of frames and aims to capture temporal relations across multiple frames by a temporal relation block. Temporal relation information is vital for video classification, as it is capable of capturing high-level interactions among entities (subjects, objects, scenes, etc.) over a long temporal series, which are significant for recognizing events with complex temporal dynamics. To take advantage of this cue, we apply a 2D CNN to video frames to extract appearance features. Then, these features are fed into the temporal relation block to learn a multi-scale temporal relation bank $\bm{l}$ across arbitrary frames. 


\textbf{Fusion module} combines outputs of the two pathways to build a more discriminative representation. More specifically, it leverages a normalization-like pipeline in which the temporal relations are transformed to two modulation parameters by two affine transformations, and the produced parameters $\mathcal{F}_1(\bm{l})$  $\mathcal{F}_2(\bm{l})$ multiplied and added with the holistic feature $\bm{g}$ to yield the normalized activation element-wisely. Finally, the fused feature $\bm{z}$ is obtained by concatenating the normalized activation with an additional holistic feature $\bm{g}$.

In what follows, we detail the temporal relation block and fusion module.

\subsection{Temporal Relation Block} \label{B}

The purpose of temporal relational reasoning lies in linking meaningful transformations among entities over time. ~\cite{he2019stnet} is intended to construct a fully connected graph among entities in video frames and calculate pairwise energy functions among node pairs in the graph to model temporal relations. Inspired by this work, we aim at capturing temporal relations among arbitrary frames. Instead of utilizing a fully connected graph among video frames which inevitably increases computation and redundancy, we make use of a sampling strategy to sample multiple snippets and learn relational representations using a group of multilayer perceptrons (MLPs). Note that each sampled snippet contains a variable number of frames for the purpose of learning multi-scale relational representations. 

Formally, suppose that we have extracted an appearance feature set $\mathcal{V} =\left \{ \bm {f}_1, \bm {f}_2,...,\bm {f}_N \right \}$ of video frames by a 2D CNN, where $\bm {f}_i$ denotes the 256-dimensional feature vector of the $i$-th video frame, and $N$ is the number of frames. We randomly sample \emph{m} vectors from $\mathcal{V}$ and concatenate them to $\bm{s}_m$, where $m$ is the total number of sampled frames, the length of vector $\bm{s}_m$ is $m \times 256$. Notably, before concatenation, we rearrange sampled vectors according to the original temporal order. The corresponding \emph{m-frame} relation function is defined as below:
\begin{equation}
    {\rm R}({\bm s}_m) =h_{\phi{_m}}({\bm s}_m)\,,
\end{equation}
where the input is the concatenated vector ${\bm s}_m= [\bm {f}_i,\ \bm {f}_j,\ ...,\ \bm {f}_p]$, $i, j, p \in [1,N] $, $m \in [2,N]$, and $[\cdot,\cdot ]$ denotes concatenation.  $h_{\phi{_m}}$ is a two-layer MLP with parameters $\phi{_m}$ and learns the transformations among \emph{m} feature vectors. The parameters of $h_{\phi{_m}}$ are learned separately with respect to each ${\bm s}_m$. With variant values \emph{m}, temporal relations at multiple timescales can be yielded and further concatenated to build a multi-scale temporal relation bank ${\bm{l}} = [{\rm R}({\bm s}_2),\ {\rm R}({\bm s}_3),\ ...,\ {\rm R}({\bm s}_N)]$.

The temporal relation block is a basic computational unit with an input feature set $\mathcal{V}$ and an output temporal relation bank $\bm l$, and can be easily plugged into any classification CNN models. Fig.~\ref{trb} illustrates the structure of our temporal relation block.


\subsection{Fusion Module} \label{C}

\begin{figure}[h]
	\centering
 	\includegraphics[width = 1\linewidth]{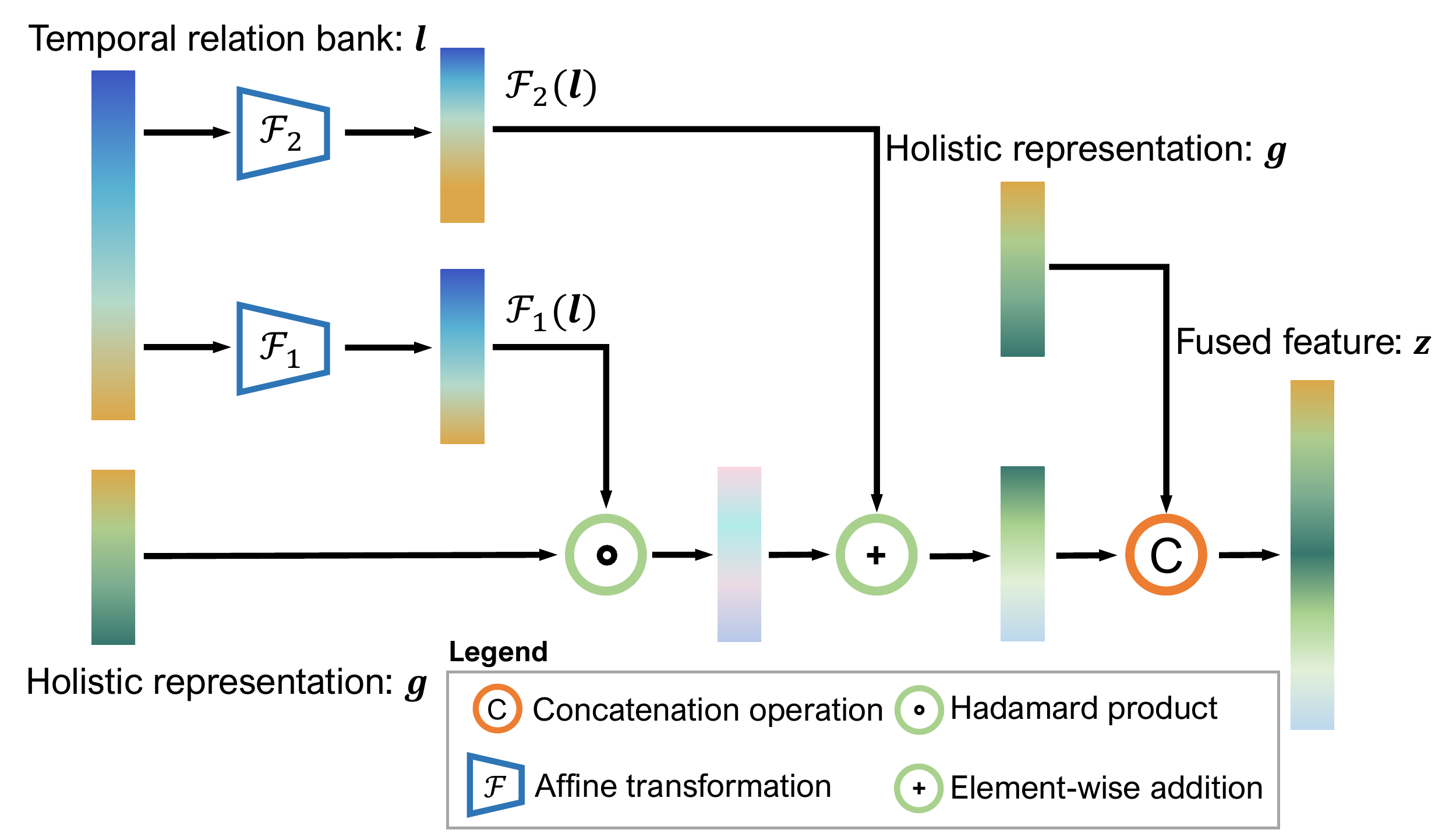}
	\caption{\textbf{Fusion module.} Two affine transformations are applied on the temporal relation bank $\bm{l}$ to produce two vectors, $\mathcal{F}_1(\bm{l})$ and $\mathcal{F}_2(\bm{l})$, respectively. Afterwards, the Hadamard product and addition operation are applied on them with $\bm{g}$. Finally,  the  output vector is concatenated with $\bm{g}$ to yield the fused feature $\bm{z}$.}
	\label{fuse}
\end{figure}

Outputs from the holistic representation pathway and temporal relation pathway are integrated by a fusion module that encodes spatiotemporal correspondences between holistic features and temporal relations. Spatiotemporally registering the two features is vital for encoding spatiotemporal correspondences. Motivated by conditional normalization~\cite{ioffe2015batch,su2020adapting}, we present a novel fusion module where the two features are spatiotemporally registered by modulating the holistic features according to temporal relations.  The multi-scale temporal information is leveraged to refine the temporal representations in holistic features. Specifically, the module utilizes a normalization-like pipeline in which the temporal relations are transformed to two modulation parameters by two affine transformations, and the produced parameters $\mathcal{F}_1(l)$ and $\mathcal{F}_2(l)$ are multiplied and added with $\bm g$ to yield the normalized activation element-wisely. Finally, the fused feature $\bm z$ is obtained by concatenating the normalized activation with an additional holistic feature $\bm g$. The fusion equation is as follows:

\begin{equation}
   {\bm z}=\left [\mathcal{F}_1({\bm l})\odot {\bm g} +\mathcal{F}_2({\bm l}),\ {\bm g} \right ] \,,
\end{equation} 
where $\mathcal{F}_1$ and $\mathcal{F}_2$ are affine transformations and aim to produce the modulation parameters, $\odot$ denotes a Hadamard production, and $[\cdot,\cdot ]$ denotes concatenation. The overall structure of fusion module is illustrated in Fig.~\ref{fuse}. We concatenate an additional holistic feature $\bm g$ with the modulated feature to yield the final fused feature $\bm z$. This is for enriching the spatial information that is important for distinguishing events with simple dynamics. For validating its effectiveness, We further compare it with several existing fusion methods in ablation study (See Section~\ref{Ablation_Experiments}).

\subsection{Implementation Details} \label{D}
In this subsection, we describe the implementation of our FuTH-Net.


\textbf{Holistic representation pathway.} We convert a typical image classification architecture, Inception-v1~\cite{szegedy2015going}, into a 3D architecture by inflating all convolutions and pooling filters. The 3D convolutions are created by endowing 2D ones with an additional temporal dimension. Furthermore, we would like to bootstrap the 2D network weights pretrained on ImageNet into the 3D model. To achieve this, the 3D model could be implicitly pretrained on ImageNet by converting images into fixed videos. We replicate weights of 2D convolutions $N$ times along the temporal dimension and then divide them by $N$ to produce pretrained parameters for the 3D model. Moreover, we optimize hyperparameters for convolutions and pooling operations (e.g., stride and pooling size) to effectively capture representative temporal dynamics. In detail, we use $1 \times 3 \times 3 $ kernels with $1\times2\times2$ strides in the first two max pooling layers for remaining initial temporal information. The final average pooling layer exploits a $2 \times 7 \times 7$ kernel to produce a 1024-dimension feature vector which is regarded as the holistic representation
$\bm g$.

\begin{figure}[]
	\centering
 	\includegraphics[width = 1\linewidth]{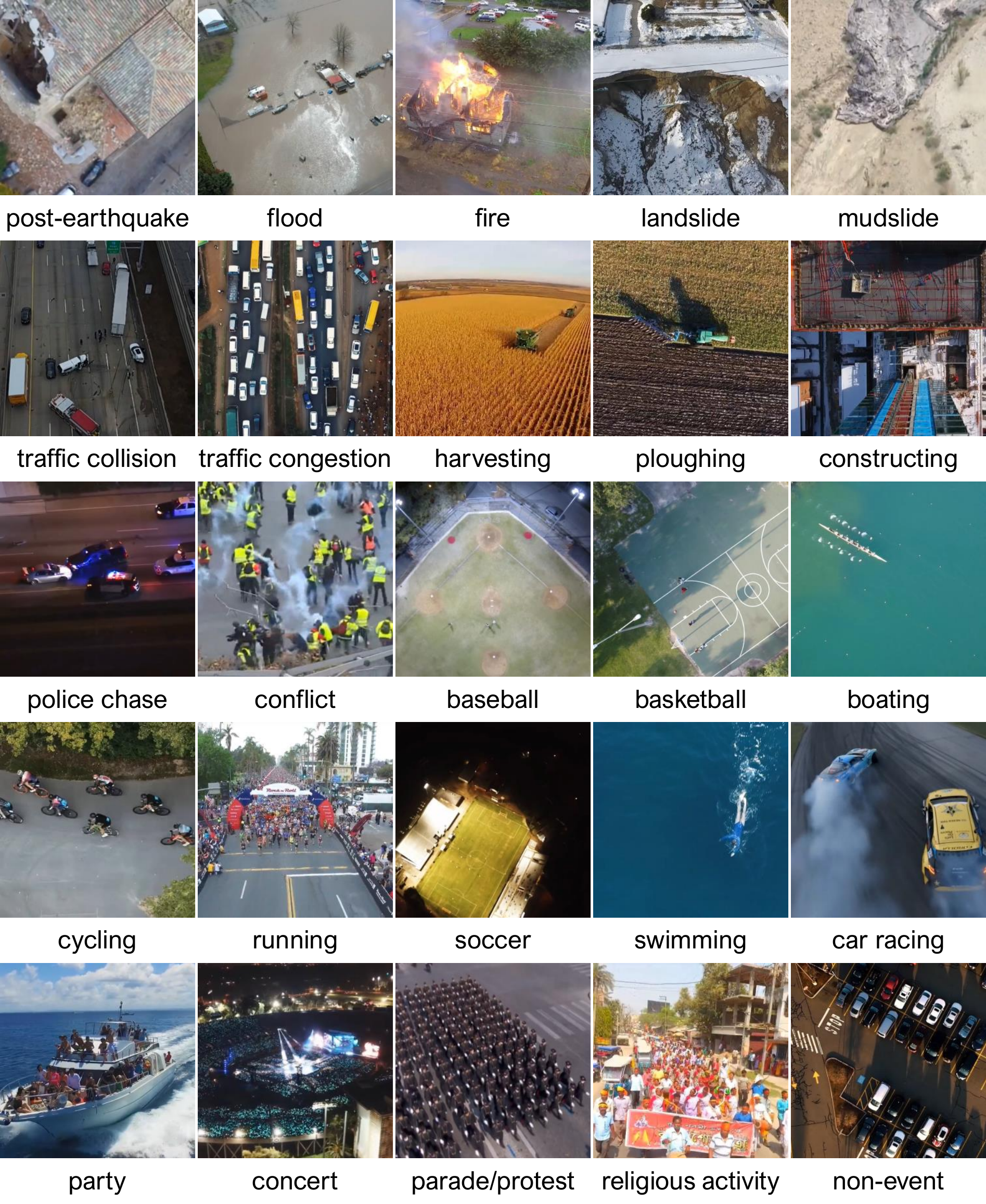}
	\caption{\textbf{Overview of the ERA dataset.} We show the middle frame of one video in each class.}
	\label{era_samples}
\end{figure}

\textbf{Temporal relation pathway.} We utilize Inception-v1 with batch normalization pretrained on ImageNet as our feature extraction model to generate a 1024-dimension feature vector for each frame. Subsequently, a feature bank with the size of $n\times1024$ for an input video is produced, where $n$ is the number of input video frames. Moreover, $\phi_m$ is a two-layer MLP with 256 units, and each layer is followed by a batch normalization~\cite{ioffe2015batch} layer and a ReLU activation function. $(N-1)$ temporal relations are extracted by $\phi_m$ and then concatenated into the final multi-scale temporal relation bank with the dimension of $256\times (N-1)$. The number of input frames is set to 16 in both two pathways.



\textbf{Fusion module.} Two simple MLP with dropout operations are exploited to implement the two affine transformations which are employed on $\bm l$ to yield two 1024-dimension vectors, $\mathcal{F}_1 (\bm l)$ and $\mathcal{F}_2(\bm l)$. The final fused feature is a 2048-dimension vector.

\begin{figure}[h!]
	\centering
 	\includegraphics[width = 1\linewidth]{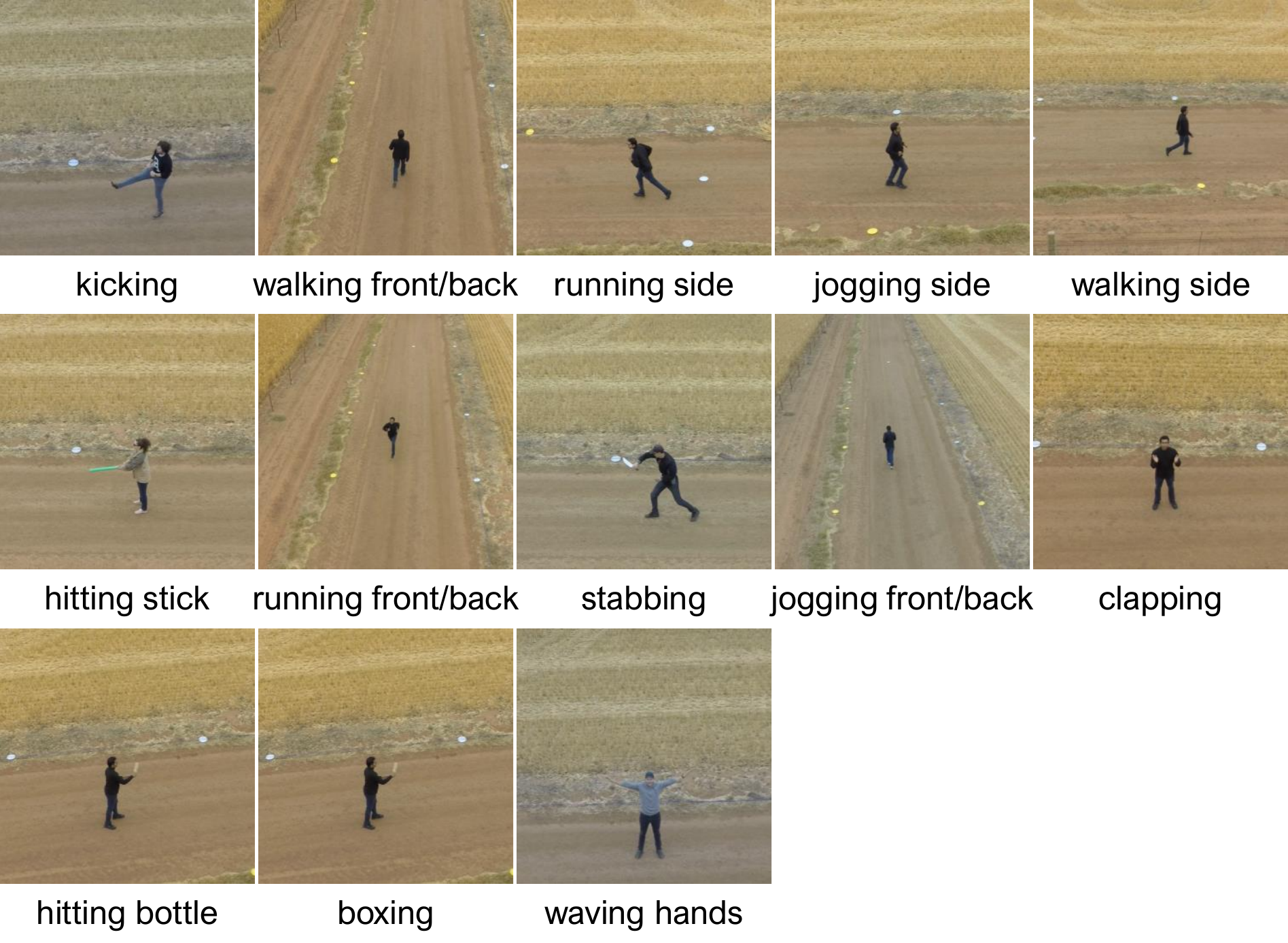}
	\caption{\textbf{Overview of the Drone-Action dataset.} We show the middle frame of one video in each class.}
	\label{drone_samples}
\end{figure}

\textbf{Training schedule.} The network is trained on PyTorch\footnote[1]{\url{https://pytorch.org/}} framework and runs on one NVIDIA Tesla P100 GPU\footnote[2]{\url{https://www.nvidia.com/en-us/data-center/tesla-p100/}} with 16 GB on-board memory. We train our model with a stochastic gradient descent (SGD)~\cite{bottou2010large} optimizer using a momentum of $0.9$ and a weight decay of $0.0005$. Due to the limitation of GPU memory, we utilize a multi-stage training strategy. Specifically, the whole training procedure is composed of three phases. First, we train the holistic representation pathway for $100$ epochs with a batch size of $6$ and a learning rate of $0.001$. Then, we train the temporal relation pathway with a learning rate of $0.0001$ and the same epochs and batch size, while keeping weights of the holistic representation pathway fixed. Finally, the fusion module is trained for $120$ epochs with weights of two pathways fixed.

\section{Experiments} \label{III}

\begin{table}[]
\caption{\textbf{Dataset overview.} We provide variable details of the two datasets.}
\label{dataset_details}
\centering
\begin{tabular}{@{}c|cc@{}}
\toprule
\textbf{}              & \textbf{ERA Dataset\cite{mou2020era}}               & \textbf{Drone-Action Dataset\cite{perera2019drone} }  \\
\midrule
\midrule
Type of Task      & general event recognition & human  action recognition \\ [4pt]
Data Source            & YouTube                   & self-collected (actor staged) \\ [4pt]
\# Classes & 25                        & 13                            \\ [4pt]
Video Size        & 640 $\times$ 640          & 1920 $\times$ 1080  \\ [4pt]
Video Duration    & 5s                        & 5s $\sim$ 21s                  \\ [4pt]
\# Samples           & 2864                      & 240                           \\
\bottomrule
\end{tabular}
\end{table}

In this section, we first introduce aerial video recognition datasets, competitors, and evaluation metrics in Section \ref{Experimental Setup}. Then, we perform ablation studies to investigate the complementarity between the holistic representation pathway and temporal relation pathway as well as the effectiveness of our fusion module in Section \ref{Ablation_Experiments}. Furthermore, we assess the performance of our FuTH-Net on two different aerial video recognition datasets, ERA and Drone-Action, and analyze experimental results in Section \ref{Experiment_A} and \ref{Experiment_B}, respectively.

\subsection{Experimental Setup} \label{Experimental Setup}

\begin{table*}[]
\footnotesize
\caption{\textbf{Comparison with state-of-the-art methods on the ERA dataset.} We show the per-class precision and overall accuracy on the test set. The best precision/accuracy is shown in \textbf{bold}.}
\label{tab:acc_era}
\centering
\begin{adjustwidth}{-0.2cm}{0cm}
\begin{threeparttable}
\begin{tabular}{p{1.75cm}|*{25}{p{0.17cm}<{\centering}}p{0.2cm}<{\centering}p{0.2cm}<{\centering}}
\toprule
\textbf{Model} & \rotatebox{75}{post-earthquake} & \rotatebox{75}{flood} & \rotatebox{75}{fire} & \rotatebox{75}{landslide} & \rotatebox{75}{mudslide} & \rotatebox{75}{traffic collision} & \rotatebox{75}{traffic congestion} & \rotatebox{75}{harvesting} & \rotatebox{75}{ploughing} & \rotatebox{75}{constructing} & \rotatebox{75}{police chase} & \rotatebox{75}{conflict} & \rotatebox{75}{baseball} & \rotatebox{75}{basketball} & \rotatebox{75}{boating} & \rotatebox{75}{cycling} & \rotatebox{75}{running} & \rotatebox{75}{soccer} & \rotatebox{75}{swimming} & \rotatebox{75}{car racing} & \rotatebox{75}{party} & \rotatebox{75}{concert} & \rotatebox{75}{parade/protest} & \rotatebox{75}{religious activity} & \rotatebox{75}{non-event} & \textbf{OA} & $\kappa$\\
\midrule
\midrule
C3D$^\dag$ & 23.1 & 24.3 & 30.9 & 19.5 & 32.9 & 7.00 & 15.5 & 27.5 & 36.1 & 45.5 & 50.0 & 18.2& 40.9 & 37.0 & 47.5 & 20.6 & 12.0 & 58.3 & 36.2 & 16.7 & 25.8 & 38.2 & 37.8 & 27.5 & 29.6 & 30.4 & 0.21\\
C3D$^\ddag$ & 27.9 & 56.5 & 32.7 & 10.2 & 23.9 & 8.30 & 38.5 & 42.3 & 31.1 & 40.0 & 51.9 & 11.1 & 45.7 & 48.9 & 41.9 & 13.6 & 9.30 & 41.9 & 38.2 & 18.2 & 17.4 & 32.0 & 28.1 & 35.8 & 28.5 & 31.1 & 0.23\\
P3D$^\dag$-{\tiny ResNet-199} & 43.6 & 65.9 & 66.7 & 35.5 & 48.7 & 20.0 & 37.8 & 77.4 & 70.8 & 62.0 & \textbf{81.6} & 22.2& 66.7 & 63.1 & 55.4 & 35.6 & 35.3 & 76.2 & 57.4 & 40.0 & 54.5 & 37.5 & 38.7 & 47.8 & 37.4 & 50.7  & 0.47\\
P3D$^\ddag$-{\tiny ResNet-199} & 72.4 & 76.3 & 84.8 & 24.5 & 38.2 & 35.6 & 40.8 & 56.9 & 67.4 & 71.4 & 57.9 & 50.0 & 70.4 & 78.8 & 71.7 & 47.1 & 60.0 & 79.5 & 68.1 & 40.9 & 59.1 & 37.0 & 49.1 & 55.9 & 37.9 & 53.3  & 0.51\\
I3D$^\dag$-{\tiny Inception-v1} & 40.4 & 63.5 & 68.9 & 22.6 & 46.3 & 17.6 & 55.0 & 61.5 & 50.0 & 53.3 & 73.2 & 50.0 & 75.0 & 69.4 & 60.7 & 61.9 & 53.3 & 70.8 & 52.5 & 50.0 & 57.1 & 50.7 & 40.3 & 49.0 & 35.8 & 51.3 & 0.48\\
I3D$^\ddag$-{\tiny Inception-v1} & 60.0 & 68.1 & 65.7 & 29.0 & 60.4 & \textbf{51.5} & 52.2 & 67.1 & 66.7 & 54.2 & 64.8 & 57.9 & \textbf{85.0} & 61.9 & \textbf{86.4} & 75.0 & 44.4 & 77.6 & 64.1 & 65.2 & 53.7 & 50.0 & 47.8 & 65.1 & 43.0 & 58.5 & 0.55\\
TRN$^\dag$-{\tiny BNInception} & \textbf{84.8} & 71.4 & 82.5 & 51.2 & 50.0 & 46.8 & \textbf{66.7} & 68.1 & 77.4 & 52.4 & 70.5 & 75.0 & 64.5 & 67.7 & 84.0 & 56.1 & 55.2 & \textbf{83.3} & 72.9 & 61.1 & 62.0 & 48.9 & 44.6 & 62.8 & 51.1 & 62.0 & 0.58\\

TRN$^\ddag$-{\tiny Inception-v3} & 69.2 & 87.8 & 88.9 & \textbf{65.8} & 60.0 & 44.1 & 58.3 &  78.1 & \textbf{90.7} & 70.8 & 73.3 & 28.6 & 83.3 & 72.7 & 73.7 & 60.0 & \textbf{66.7} & 73.6 & 70.6 & 63.6 & 65.1 & 47.7 & 42.7 & 65.1 & 47.9 & 64.3 & 0.60\\



SlowFast$^\dag$ & 70.1 & \textbf{88.0} & 83.3 & 57.2 & 67.3 & 51.4 & 56.2 & 68.4 & 87.6 & \textbf{82.0} & 75.1 & \textbf{75.5} & 40.8 & 70.3 & 71.8 & 61.4 & 54.7 & 78.2 & 72.9 & 74.3 & 50.4 & 70.3 & 50.6 & \textbf{65.7} & 60.7 & 64.9 & \textbf{0.63}\\

Multigrid$^\dag$ & 69.8 & 71.7 & \textbf{89.5} & 54.7 & 64.1 & 47.4 & 59.4 & \textbf{78.4} & 73.4 & 69.4 & 72.4 & 51.8 & 63.8 & 74.7 & 76.2 & \textbf{75.2} & 52.1 & 71.1 & 69.6 & 67.5 & \textbf{66.1} & 74.4 & 55.7 & 62.3 & 57.4 & 65.3 & 0.62\\

\hline

FuTH-Net & 72.7 & 75.5 & 87.5 & 57.1 & \textbf{74.5} & 34.0 & 56.0 & 76.6 & 71.2 & 81.4 & 76.5 & 36.0 & 78.0 & \textbf{85.4} & 80.4 & 73.6 & 16.3 & 64.5 & \textbf{80.4} & \textbf{84.2} & 56.0 & \textbf{89.8} & \textbf{65.3} & 63.0 & \textbf{63.9} & \textbf{66.8} & \textbf{0.63}\Tstrut\\

\bottomrule
\end{tabular}

\begin{tablenotes}
    \item[1] C3D$^\dag$ uses pre-trained weights on the Sport1M dataset as initialization; C3D$^\ddag$ uses pre-trained weights on the UCF101 dataset as initialization.
    \item[2] P3D$^\dag$-{\tiny ResNet-199} uses pre-trained weights on the Kinetics dataset as initialization; P3D$^\ddag$-{\tiny ResNet-199} uses pre-trained weights on the Kinetics-600 dataset as initialization.
    \item[3] I3D$^\dag$-{\tiny Inception-v1} uses pre-trained weights on the Kinetics dataset as initialization; I3D$^\ddag$-{\tiny Inception-v1} uses pre-trained weights on Kinetics+ImageNet as initialization.
    \item[4] TRN$^\dag$-{\tiny BNInception} uses pre-trained weights on the Something-Something V2 dataset as initialization; TRN$^\ddag$-{\tiny Inception-v3} uses pre-trained weights on the Moments in Time dataset as initialization.
    \item[5] SlowFast$^\dag$ is trained from random initialization, without using pre-training.
    \item[6] Multigrid$^\dag$ use ImageNet-pre-trained for 3D convolutions inflated from 2D convolutions following common practice.
\end{tablenotes}
\end{threeparttable}
\end{adjustwidth}
\end{table*}

\textbf{Datasets.} To evaluate the performance of FuTH-Net, we conduct experiments on two aerial video recognition datasets with standard evaluation protocols. Firstly, we use the ERA dataset\cite{mou2020era} which is an event recognition dataset and consists of $2864$ aerial event videos collected from YouTube. In this dataset, $25$ events are defined, including \emph{post-earthquake}, \emph{flood}, \emph{fire}, \emph{landslide}, \emph{mudslide}, \emph{traffic collision}, \emph{traffic congestion}, \emph{harvesting}, \emph{ploughing}, \emph{constructing}, \emph{police chase}, \emph{conflict}, \emph{baseball}, \emph{basketball}, \emph{boating}, \emph{cycling}, \emph{running}, \emph{soccer}, \emph{swimming}, \emph{car racing}, \emph{party}, \emph{concert}, \emph{parade/protest}, \emph{religious activity}, and \emph{non-event} (see Fig. \ref{era_samples}). Then, the Drone-Action dataset\cite{perera2019drone} for human action classification in aerial videos is utilized to further assess the performance of models. In this dataset, $240$ self-taken aerial videos are collected, and $13$ different actions are defined: \emph{kicking}, \emph{walking front/back}, \emph{running side}, \emph{jogging side}, \emph{walking side}, \emph{hitting stick}, \emph{running front/back}, \emph{stabbing}, \emph{jogging front/back}, \emph{clapping}, \emph{hitting bottle}, \emph{boxing}, and \emph{waving hands} (see Fig. \ref{drone_samples}). Table \ref{dataset_details} exhibits details of the two datasets.


In the preprocessing phase, we transform video clips of the Drone-Action dataset into the same data structure as the ERA dataset. Since durations of videos in the Drone-Action dataset range from 5 to 21 seconds, we cut them to 5-second clips. Afterwards, each frame is cropped and resized to a size of $640 \times 640$. For both datasets, we sample 16 frames from each video clip with a fixed sampling rate.

\textbf{Competitors.} We compare the proposed network with several state-of-the-art video classification models.

\begin{itemize}
    \item
    C3D~\cite{tran2015learning}. C3D (3D convolutional network) aims to extract spatiotemporal features with 3D convolutional filters and pooling layers. Compared to conventional 2D CNNs, 3D convolutions and pooling operations in C3D can preserve the temporal information of input signals and model motion as well as appearance simultaneously. Moreover, authors in \cite{tran2015learning} demonstrate that the optimal size of 3D convolutional filters is 3$\times$3$\times$3. In our experiments, we test two C3D\footnote[3]{\url{https://github.com/tqvinhcs/C3D-tensorflow}} networks with pre-trained weights on the Sport1M dataset~\cite{KarpathyCVPR14} and the UCF101 dataset~\cite{soomro2012ucf101} (see C3D$^\dag$ and C3D$^\ddag$ in Table~\ref{tab:acc_era}), respectively.
    \item
    P3D ResNet~\cite{qiu2017learning}. P3D ResNet (pseudo-3D residual network) is composed of pseudo-3D convolutions, where conventional 3D convolutions are decoupled into 2D and 1D convolutions in order to learn spatial and temporal information separately. With such convolutions, the model size of a network can be significantly reduced, and the utilization of pre-trained 2D CNNs is feasible. Besides, inspired by the success of ResNet~\cite{he2016deep}, P3D ResNet employs ResNet-like architectures to learn residuals in both spatial and temporal domains. In our experiments, we test two 199-layer P3D ResNet\footnote[4]{\url{https://github.com/zzy123abc/p3d}} (P3D-{\tiny ResNet-199}) models with pre-trained weights on the Kinetics dataset~\cite{kinetics400} and the Kinetics-600 dataset~\cite{kinetics600} (see P3D$^\dag$-{\tiny ResNet-199} and P3D$^\ddag$-{\tiny ResNet-199} in Table~\ref{tab:acc_era}), respectively.
    \item
    \begin{figure}[!h]
	\centering
 	\includegraphics[width = 1\linewidth]{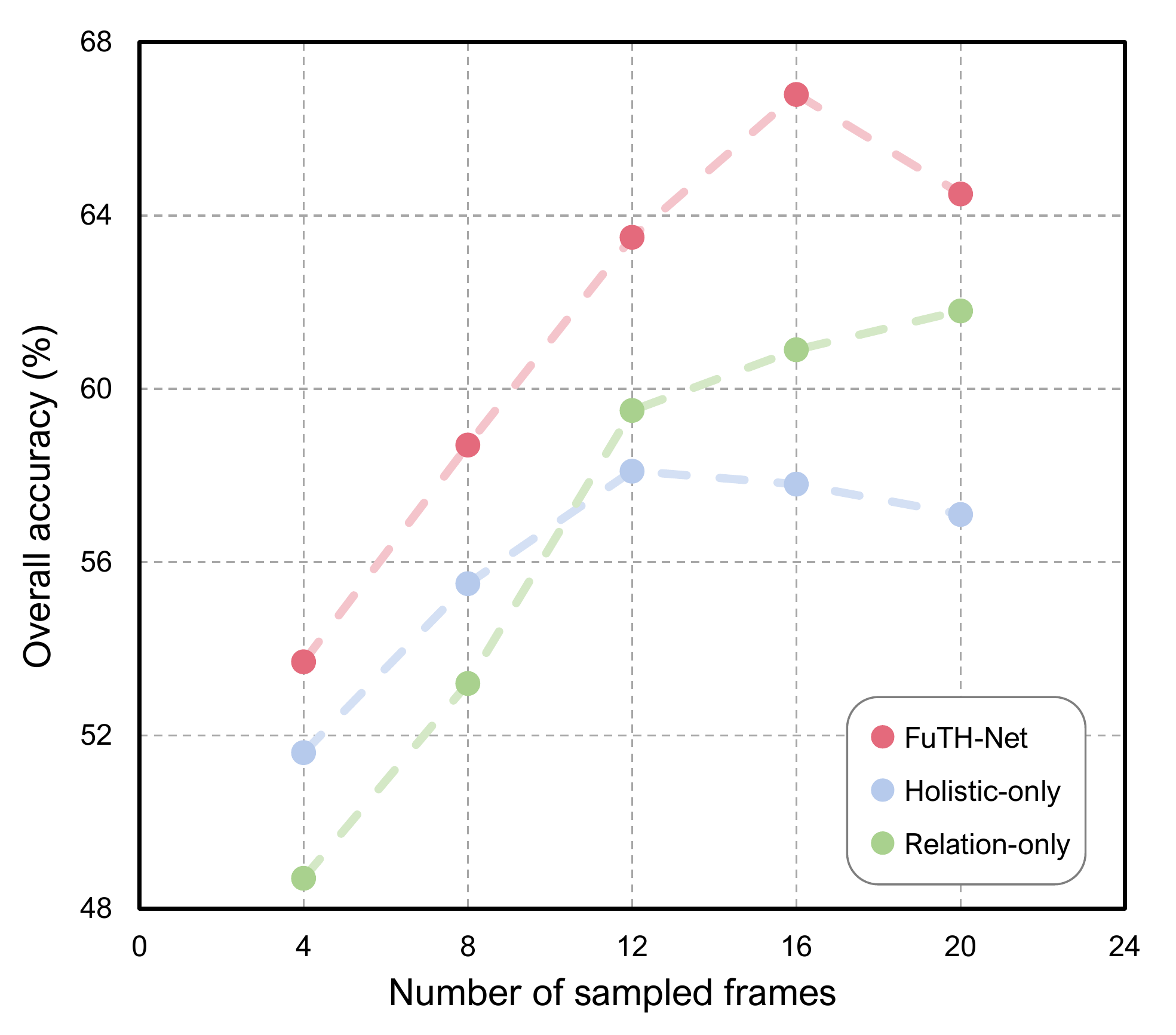}
	\caption{\textbf{FuTH-Net \emph{vs.} Holistic-only \emph{vs.} Relation-only.} Overall accuracies of FuTH-Net (\textcolor[RGB]{228,106,124}{\textbf{red}}), Holistic-only (\textcolor[RGB]{182,202,236}{\textbf{blue}}), and Relation-only (\textcolor[RGB]{169,209,142}{\textbf{green}}) with different numbers of sampled frames on the ERA dataset.}
	\label{LQ-only}
\end{figure}
    I3D~\cite{carreira2017quo}. I3D (inflated 3D ConvNet) expands 2D convolution and pooling filters to 3D, which are then initialized with inflated pre-trained models. Particularly, weights of 2D networks pre-trained on the ImageNet dataset are replicated along the temporal dimension. With this design, not only 2D network architectures but also pre-trained 2D models can be efficiently employed to increase the learning efficiency and performance of 3D networks. To assess the performance of I3D on our dataset, we test two I3D\footnote[5]{\url{https://github.com/LossNAN/I3D-Tensorflow}} models whose backbones are both Inception-v1~\cite{szegedy2015going} (I3D-{\tiny Inception-v1}) with pre-trained weights on the Kinetics dataset~\cite{kinetics400} and Kinetics+ImageNet, respectively (see I3D$^\dag$-{\tiny Inception-v1} and I3D$^\ddag$-{\tiny Inception-v1} in Table~\ref{tab:acc_era}). 
    \item
    TRN~\cite{zhou2018temporal}. Temporal relation network (TRN) is proposed to recognize human actions by reasoning about multi-scale temporal relations among video frames. By leveraging the proposed plug-and-play relational reasoning module, TRN can even accurately predict human gestures and human-object interactions through sparsely sampled frames. For our experiments, we test TRNs\footnote[6]{\url{https://github.com/metalbubble/TRN-pytorch}} with 16 multi-scale relations and select the Inception architecture as the backbone. Notably, we experiment two variants of the Inception architecture: BNInception~\cite{DBLP:conf/icml/IoffeS15} and Inception-v3~\cite{inceptionv3}. We initialize the former with weights pre-trained on the Something-Something V2 dataset~\cite{DBLP:conf/iccv/GoyalKMMWKHFYMH17} (TRN$^\dag$-{\tiny BNInception} in Table~\ref{tab:acc_era}) and the latter with weights pre-trained on the Moments in Time dataset~\cite{moments} (TRN$^\ddag$-{\tiny Inception-v3} in Table~\ref{tab:acc_era}).
    \item
    SlowFast~\cite{feichtenhofer2019slowfast}. SlowFast network is a two-pathway architecture in which a Slow pathway is designed for operating at low frame rate to capture spatial semantic information, and a Fast pathway aims at operating at high frame rate to learn motion at fine temporal resolution. To assess the performance of SlowFast on our dataset, we test one SlowFast\footnote[7]{\url{https://github.com/facebookresearch/SlowFast}} model (see SlowFast$\dag$ in Table~\ref{tab:acc_era}) whose backbone is ResNet~\cite{he2016deep} without pre-training.
    \item
    Multigrid~\cite{wu2020multigrid}. Multigrid training method utilizes variable mini-batch shapes with different spatiotemporal resolutions in the training phase. The different shapes are generated by resampling the training data on multiple sampling grids. The novel training method yields a significant out-of-the-box training speedup for different models (I3D, SlowFast). In our experiments, we use this training method test SlowFast network\footnote[8]{\url{https://github.com/facebookresearch/SlowFast/tree/master/projects/multigrid}} (see Multigrid$\dag$ in Table~\ref{tab:acc_era}) with ImageNet-pre-training.

\end{itemize}

\textbf{Evaluation metrics.} We make use of the per-class precision, overall accuracy, confusion matrix and kappa coefficient as evaluation metrics for comparing the performance of different models. Specifically, the pre-class precision is calculated with the following equation:

\begin{equation}
    precision = \frac{true \: positives}{true \: positives + false \: positives}\,.
\end{equation}

\begin{table}[]
\caption{\textbf{Comparison with different hybrid models.} We compare our FuTH-Net with different hybrid models using different fusion methods on ERA and Drone-Action datasets.}
\label{hybrid_models}
\centering

\begin{tabular}{@{}c|cc|cc@{}}
\toprule
\multirow{2}{*}{Model} & \multicolumn{2}{c|}{\textbf{ERA Dataset}}                                       & \multicolumn{2}{c}{\textbf{Drone-Action Dataset}}                              \\ [4pt] \cline{2-5} \Tstrut
                       & \multicolumn{1}{l}{Concatenation} & \multicolumn{1}{l|}{Ours} & \multicolumn{1}{l}{Concatenation} & \multicolumn{1}{l}{Ours} \\ 
\midrule
\midrule


C3D+TRN & 46.7 & 45.3 & 58.4 & 60.1 \\ [4pt] P3D+TRN & 51.9                             & 52.4                              &  82.9                             & 86.6                             \\ [4pt]
I3D+TRN                &  58.4                              & 60.8                               &  84.3                             & 85.2                             \\ [4pt]
FuTH-Net             &  \textbf{64.8}                             & \textbf{66.8}                               & \textbf{87.7}                              & \textbf{88.4}     \\
\bottomrule
\end{tabular}
\end{table}

The overall accuracy is computed by dividing the number of correctly classified test samples with that of all test samples. Moreover, the confusion matrix is visualized to illustrate the classification performance of variant models. Each element of the matrix denotes the number of instances that belong to the ground-truth class (X-axis) but are classified as the predicted class (Y-axis). For an explicit visualization, we normalize the confusion matrix by dividing each element with the sum of each row. In addition, the kappa coefficient is leveraged to evaluate consistency and classification precision. It considers both the overall accuracy and the variations in the number of samples in each category.

\subsection{Ablation Studies} \label{Ablation_Experiments}

To evaluate the complementarity between two pathways and effectiveness of the fusion module, we conduct ablation studies on the ERA and Drone-Action datasets.
\begin{table*}[t!]
\footnotesize
\caption{\textbf{Comparison with state-of-the-art methods on the Drone-Action dataset.} We show the per-class precision and overall accuracy on the test set. The best precision/accuracy is shown in \textbf{bold}}.
\label{tab:acc_drone}
\centering
\begin{tabular}{p{1.75cm}|*{13}{p{0.65cm}<{\centering}}p{0.5cm}p{0.5cm}}
\toprule
\textbf{Model} & \rotatebox{65}{kicking} & \rotatebox{65}{walking front/back} & \rotatebox{65}{running side} & \rotatebox{65}{jogging side} & \rotatebox{65}{walking side} & \rotatebox{65}{hitting stick} & \rotatebox{65}{running front/back} & \rotatebox{65}{stabbing} & \rotatebox{65}{jogging front/back} & \rotatebox{65}{clapping} & \rotatebox{65}{hitting bottle} & \rotatebox{65}{boxing} & \rotatebox{65}{waving hands} & \textbf{OA} & $\kappa$\\
\midrule
\midrule

C3D$^\dag$ & 48.3 & 61.5 & 23.5 & 77.3 & 12.0 & 31.0 & 0.71 & 0.34 & 47.4 & 28.6 & 24.1 & 29.6 & 00.0 & 31.6 & 0.25\\

C3D$^\ddag$ & 31.0 & 80.8 & 35.3 & 13.6 & 48.0 & 24.1 & 21.4 & 0.69 & 42.1 & 42.9 & 37.9 & 0.37 & 00.0 & 30.3 & 0.24\\
P3D$^\dag$-{\tiny ResNet-199} & \textbf{100} & 73.1 & 41.2 & 86.4 & 96.0 & \textbf{100} & 57.1 & 93.1 & 36.8 & 78.6 & 93.1 & 85.2 & \textbf{100} & 83.0 & 0.81\\
P3D$^\ddag$-{\tiny ResNet-199} & \textbf{100} & 69.2 & 47.1 & 72.7 & 96.0 & \textbf{100} & 50.0 & 82.8 & 52.6 & 85.7 & 93.1 & 88.9 & \textbf{100} & 82.3 & 0.81\\

I3D$^\dag$-{\tiny Inception-v1} & 78.3 & 70.4 & 28.6 & 17.9 & 84.2 & 15.2 & 47.1 & 13.6 & 50.0 & 75.0 & 64.1 & 60.0 & 60.0 & 50.7 & 0.79\\

I3D$^\ddag$-{\tiny Inception-v1} & \textbf{100} & 88.9 & 42.1 & 70.6 & \textbf{100} & \textbf{100} & \textbf{64.3} & \textbf{100} & 20.0 & 90.9 & \textbf{100} & 93.8 & \textbf{100} & 85.5 & 0.84\\

TRN$^\dag$-{\tiny BNInception} & 96.6 & 61.5 & 41.2 & 68.2 & 96.0 & \textbf{100} & 35.7 & 96.6 & 47.4 & \textbf{100} & 82.8 & 88.9 & \textbf{100} & 80.6 & 0.83\\
TRN$^\ddag$-{\tiny Inception-v3} & \textbf{100} & \textbf{96.2} & 52.9 & 86.4 & \textbf{100} & \textbf{100} & 28.6 & 89.7 & 42.1 & \textbf{100} & 86.2 & 85.2 & \textbf{100} & 85.0 & 0.82\\



SlowFast$^\ddag$ & \textbf{100.0} & 88.5 & 52.9 & 95.5 & 92.0 & \textbf{100.0} & 57.1 & 89.7 & \textbf{68.4} & 92.9 & 86.2 & \textbf{96.3} & 71.4 & 86.7 & 0.86\\

Multigrid$^\ddag$ & 93.1 & 92.3 & 47.1 & \textbf{100.0} & \textbf{100.0} & 86.2 & 50.0 & \textbf{100.0} & 63.2 & \textbf{100.0} & 82.8 & 88.9 & 92.9 & 86.4 & 0.85\\
\hline

FuTH-Net & \textbf{100} & \textbf{96.2} & \textbf{58.8} & 90.9 & \textbf{100} & \textbf{100} & 28.6 & 96.6 & 52.6 & \textbf{100} & 89.7 & \textbf{96.3} & \textbf{100}  & \textbf{88.4} & \textbf{0.87}\Tstrut\\
\bottomrule
\end{tabular}
\end{table*}

\begin{table}[!h]
\footnotesize
\centering
\begin{threeparttable}
\caption{\textbf{Ablation studies of the fusion module on the ERA and Drone-Action datasets.} We show overall accuracies of FuTH-Net and FuTH-Concat and compare them with Holistic-only and Relation-only networks.The best accuracies are shown in \textbf{bold}.}
\label{tab:acc_fusion}
\centering
\begin{tabular}{p{1.9cm}|p{1.7cm}<{\centering}|p{1.6cm}<{\centering}p{1.65cm}<{\centering}}
\toprule
\textbf{Model} & fusion & \textbf{ERA Dataset} & \textbf{Drone-Action} \\
\midrule
\midrule

Holistic-only\tnote{1} & - & 57.3 & 84.6 \\
Relation-only\tnote{2} & - & 60.4 &  85.0\Bstrut\\ 


\hline 
FuTH-Max & Max & 60.1 & 81.4\Tstrut\\
FuTH-Average & Average & 61.9 & 82.6\Tstrut\\
FuTH-Concat & Concatenation & 64.8 & 85.4\Tstrut\\
FuTH-Bilinear & Bilinear & 63.2 & 82.5\Tstrut\\
FuTH-Sum & Sum & 64.7 & 84.8\Tstrut\\
FuTH-2DConv & 2D conv & 65.1 & 86.6\Tstrut\\
FuTH-3DConv & 3D conv & 65.7 & 87.2\Tstrut\\
FuTH-Net & Ours & \textbf{66.8} & \textbf{88.4}\Tstrut\\

\bottomrule
\end{tabular}

\begin{tablenotes}
    \item[1] Holistic-only is the network with only the holistic representation pathway on top of the backbone.
    \item[2] Relation-only is the network with only the temporal relation pathway on top of the backbone.
\end{tablenotes}
\end{threeparttable}
\end{table}

\textbf{Complementarity.} We investigate the complementarity by comparing our FuTH-Net with its single-pathway versions on the EAR dataset. Specifically, instead of simultaneously utilizing both pathways, Holistic-only and Relation-only make use of holistic representation and temporal relation pathways, respectively. For a comprehensive study, we compare these models under variant video sampling strategies. As shown in Fig. \ref{LQ-only}, we sample 4, 8, 12, 16, and 20 frames from each video clip and show overall accuracies. It can be observed that FuTH-Net exhibits superior performance than the other two competitors under all sampling strategies. The combination of the two pathways brings in significant improvements, demonstrating that the multi-scale temporal dependencies captured by the temporal relation pathway are largely complementary with the holistic feature.

Moreover, we note that Holistic-only outperforms Relation-only when 4 or 8 frames are used, but is surpassed by Relation-only with increasing frames. The reason could be that a few frames are not enough for the learning of multi-scale temporal relations. Another interesting observation is that the performance of Holistic-only deteriorates when the number of sampled frames is larger than 12, which might result from information redundancy. This also has a negative effect on FuTH-Net and brings a decrement of $2.3\%$ with the number of sampled frames increasing from 16 to 20. At last, FuTH-Net reaches the best performance at 16 frames.

In addition, we jointly leverage holistic spatiotemporal features and multi-scale temporal relations for video classification. For validating the effectiveness of this combination, we compare our model with other hybrid models (i.e., C3D+TRN, P3D+TRN, and I3D+TRN) on two datasets using two fusion methods, concatenation and our fusion module. The numerical results are reported in Table.~\ref{hybrid_models}. We can observe that compared to other hybrid models, our FuTH-Net achieves the best performance with different fusion methods on two datasets. Moreover, we note that hybrid models with our fusion module outperform those with concatenation in general. Another interesting observation is that the three hybrid models do not achieve better performance than single models (i.e., TRN). For example, I3D+TRN with our fusion module achieves an OA of $60.8\%$, while TRN$^\ddag$-{\tiny Inception-v3} obtains an OA of $64.3\%$.


\begin{table}[]
\footnotesize
\centering
\caption{\textbf{Ablation studies on generations of the fused feature $\bm z$.} We show overall accuracies of models with different additional features on ERA and Drone-Action datasets .The best accuracies are shown in \textbf{bold}.}
\label{different_concatenation}
\centering
\begin{tabular}{p{2.4cm}|p{2cm}<{\centering}|p{2cm}<{\centering}}
\toprule
\textbf{Additional feature} & \textbf{ERA Dataset} & \textbf{Drone-Action} \\
\midrule
\midrule

None & 66.0 & 87.3 \\ [4pt]
Temporal relation $\bm l$ & 66.2 & 87.0 \\ [4pt]
Holitic feature $\bm g$ & \textbf{66.8} & \textbf{88.4} \\

\bottomrule
\end{tabular}
\end{table}

\begin{figure*}[h]
	\centering
 	\includegraphics[width = 1\linewidth]{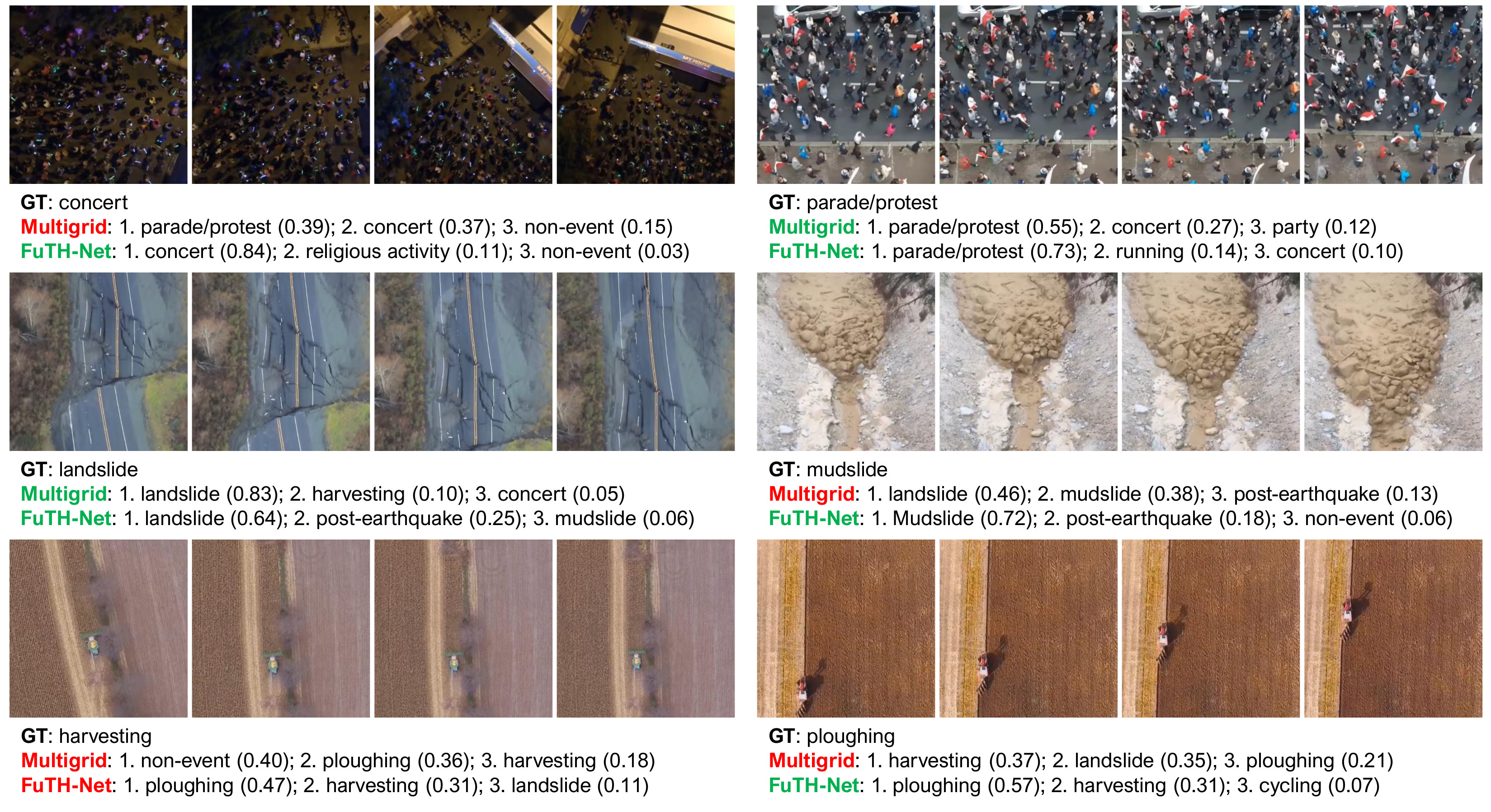}
	\caption{\textbf{Examples of predicted results on the ERA dataset.} We show results of the second best architecture, TRN, and our FuTH-Net. The ground truth label and top 3 predictions of each model are reported. Four frames are selected with 1-second interval from each example video.}
	\label{visulization_era}
\end{figure*}

\begin{figure*}[h!]
\centering
\subfigure[]{%
\includegraphics[width=1\columnwidth]{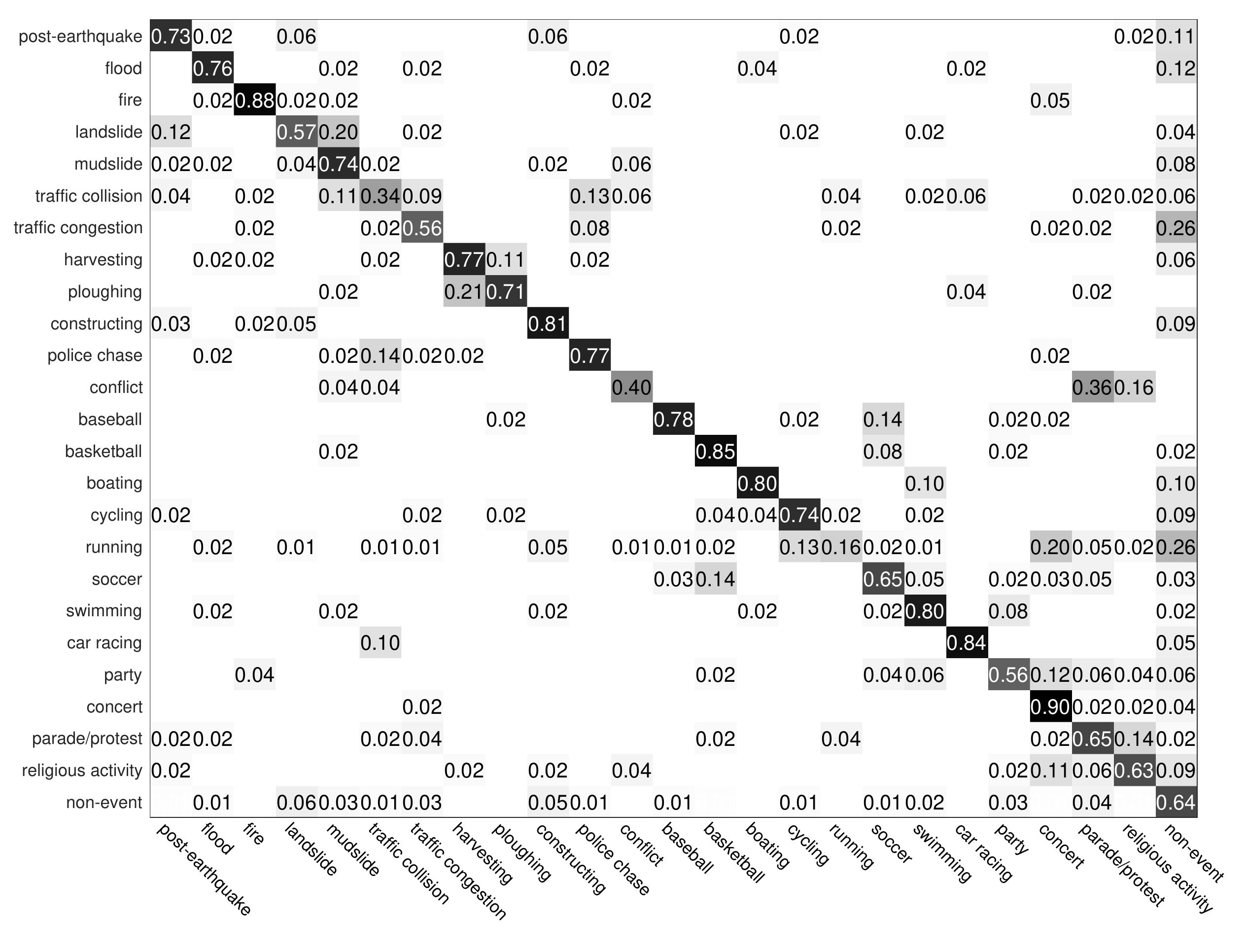}
\label{confusion_1}}
\subfigure[]{%
\includegraphics[width=1\columnwidth]{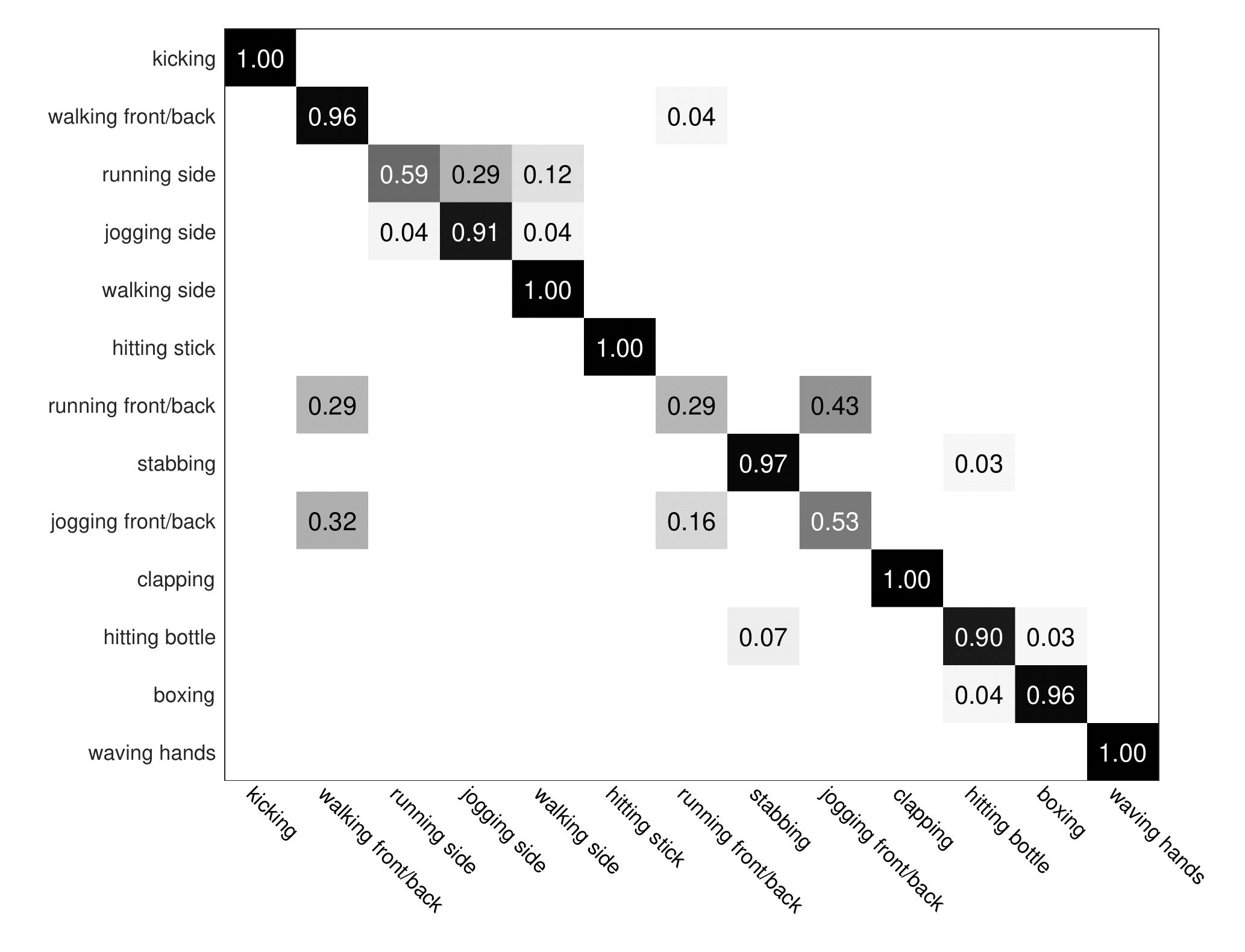}
\label{confusion_2}}
\caption{\label{fig:cm}
\textbf{Confusion matrices of the proposed network.} (a) ERA dataset; (b) Drone-Action dataset.}
\end{figure*}

\textbf{Fusion module.} As an important component in our framework, the fusion module aims to integrate features from both pathways. To validate its effectiveness, we compare the fusion module with several commonly used integration operation, such as, max, average, concatenation, bilinear, sum, 2D conv, and 3D conv. Notably, for 2D and 3D convs, the input is the concatenation of feature maps from last convolutional layers of two pathways, respectively. Table \ref{tab:acc_fusion} compares FuTH-Net to other models with different fusion modules on both the ERA and Drone-Action datasets. As can be seen in this Table, FuTH-Net provides better results than models with other different fusion methods and models with single pathways, which demonstrates that our fusion module can effectively encode high-level interactions between the two features and improve the performance.

Moreover, we concatenate an additional holistic feature $\bm g$ with the modulated feature to yield the final fused feature $\bm z$. For ablating this design, We concatenate different additional features, i.e., None and Temporal relation $\bm l$, with the modulated feature to obtain the final fused feature $\bm z$. We use these additional features to conduct ablation studies on different generations of the fused feature $\bm z$. The numerical results are reported in Table.~\ref{different_concatenation}. We can observe that the model with holistic feature $\bm g$ as the additional feature outperforms other models. Richer spatial information introduced from the holistic feature can improve the discriminant ability for events with simple dynamics.

\subsection{Results on the ERA dataset} \label{Experiment_A}

\begin{figure*}[h!]
	\centering
 	\includegraphics[width = 1\linewidth]{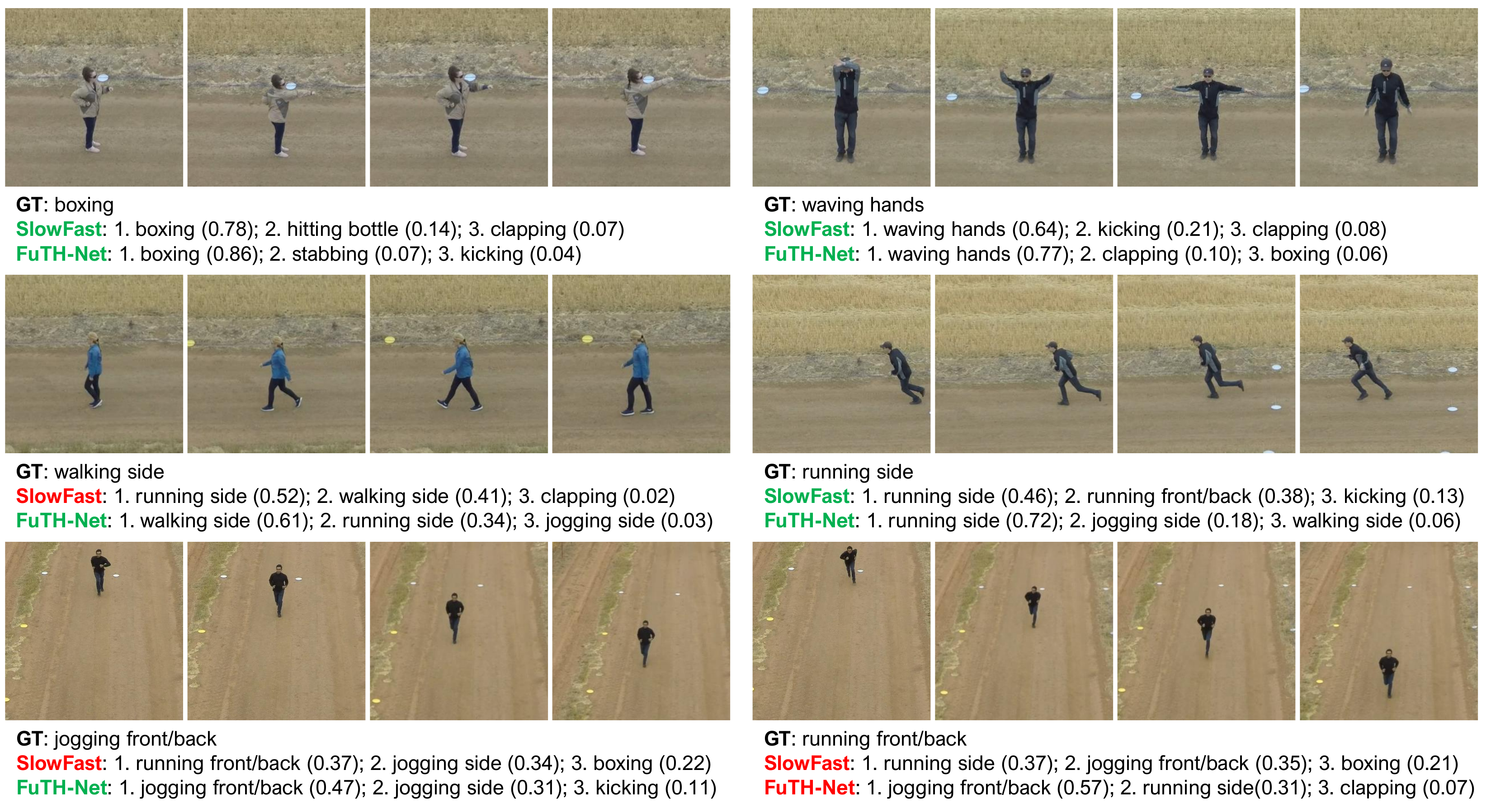}
	\caption{\textbf{Examples of predicted results on the Drone-Action dataset.} We show results of the second best architecture, TRN, and our FuTH-Net. The ground truth label and top 3 predictions of each model are reported. Four frames are selected with 1-second interval from each example video.}
	\label{visulization_drone}
\end{figure*}

We compare the proposed FuTH-Net and other competitors on the ERA dataset and report numerical results in Table \ref{tab:acc_era}. As we can see, our model has a superb performance and provides an OA of $66.8\%$ which is $1.5\%$ higher than the second best model, Multigrid$^\ddag$. And our model and Multigrid achieve the same best kappa coefficient ($0.63$). In addition, the per-class precision is also reported to evaluate the performance of different models on each class. In particular, our model achieves the highest per-class precisions for some challenging categories, such as \emph{concert} ($89.8\%$), \emph{car racing} ($84.2\%$), and \emph{parade/protest} ($65.3\%$). This is mainly because our FuTH-Net is able to capture complex dynamic information, which is crucial to distinguish events with insignificant inter-class variances. Taking \emph{concert} and \emph{parade/protest} (cf. the first row of Fig. \ref{visulization_era}) for example, they have something in common (e.g., crowd and street). However, the temporal dynamics of crowds in these two events are very different (\emph{concert}: moving randomly or standing still; \emph{parade/protest}: moving towards a certain direction). We can see that our FuTH-Net correctly predicts these two events. This also can be seen from Table \ref{tab:acc_era} that our network gains the highest precisions for these two classes, showing its effectiveness for temporal relational reasoning.

Moreover, the performance on class \emph{non-event} can reflect whether a model can distinguish specific events from normal videos. Notably, our model produces the best precision ($63.9\%$) for \emph{non-event}, which illustrates that our method is able to capture discriminative spatiotemporal features for inferring the existence of events.

Finally, the confusion matrix in Fig. \ref{confusion_1} shows more details. We can observe that some events including similar objects and scenes (e.g., "\emph{landslide vs. mudslide}"; "\emph{traffic collision vs. police chase}"; "\emph{harvesting vs. ploughing}"; "\emph{concert vs. party}") tend to be misclassified. Other competitors also suffer from this problem. Fig. \ref{visulization_era} shows some predictions of FuTH-Net and the second best model (i.e., Multigrid). It can be observed that there are a lot of visual similarities existing in textures, objects, and scenes of these events.


\subsection{Results on the Drone-Action dataset} \label{Experiment_B}


This subsection compares FuTH-Net and state-of-the-art methods on the Drone-Action dataset, and quantitative results are reported in Table \ref{tab:acc_drone}. Our FuTH-Net achieves the highest OA, $88.4\%$, and compared to SlowFast that is the second best model, an increment of $1.7\%$ can be obtained. Moreover, our model achieves the best kappa coefficient ($0.87$).

Besides, it is interesting to note that FuTH-Net shows good performance in recognizing actions in which effectively sensing motion speeds is crucial for a successful prediction. For instance, the proposed network gains the highest precisions for \emph{walking side} ($100.0\%$), \emph{running side} ($58.8\%$), and \emph{jogging side} ($100.0\%$). To further illustrate this, we show some predictions of FuTH-Net and the second best model (i.e., SlowFast) in Fig. \ref{visulization_drone}. As can be observed, the motion speeds of \emph{walking side} and \emph{running side} are variant, and our FuTH-Net succeeds in identifying them with high confidences. The bottom right example shows that \emph{running front/back} is misclassified by both FuTH-Net and SlowFast, owing to that human poses and motion speeds are very similar in this angle of view. Furthermore, the confusion matrix of the proposed network on the Drone-Action dataset shown in Fig. \ref{confusion_2} also suggests that \emph{running front/back} is easily misidentified as \emph{jogging front/back}.


\section{Conclusion} \label{IV}

In this paper, a novel method is proposed to learn feature representations from aerial videos using a two-pathway network, termed as FuTH-Net. Specifically, the proposed network exploits inflated 3D conclusions to capture a holistic feature on a holistic representation pathway. Simultaneously, a temporal relation block learns temporal relations across multiple frames on a temporal relation pathway. A novel fusion module is applied to fuse outputs from the two pathways for producing a more discriminative video representation. Furthermore, we conduct extensive experiments on two aerial video recognition datasets, ERA and Drone-Action. On the one hand, we perform ablation studies to validate the complementarity between the two pathways as well as the effectiveness of the proposed fusion module. On the other hand, we compare our model with other state-of-the-art methods. Experimental results demonstrate that the introduction of the temporal relation pathway can enhance the ability of capturing representative temporal relations. Besides, our fusion module is capable of learning high-level interactions between the holistic features and temporal relations to further boost the performance. The outstanding performance on the two datasets further illustrates the superior capability of FuTH-Net for remote sensing video recognition and its powerful generalization capability across different tasks (event classification and human action recognition).


%





\ifCLASSOPTIONcaptionsoff
  \newpage
\fi

\bibliographystyle{IEEEtran}
\bibliography{IEEEabrv, references}

\end{document}